\documentclass[runningheads]{llncs}
\usepackage{graphicx}

\usepackage{tikz}
\usepackage{comment}
\usepackage{amsmath,amssymb} 
\usepackage{color}

\usepackage[accsupp]{axessibility}  

\usepackage{graphicx}
\graphicspath{{assets/}} 
\usepackage{amsmath}
\usepackage{amssymb}
\usepackage{booktabs}
\usepackage{subfigure}
\usepackage{framed}

\usepackage{fancyref}
\usepackage{cite}
\DeclareMathAlphabet\mathcalbf{OMS}{cmsy}{b}{n}
\usepackage{mathabx}
\usepackage{mathtools}
\usepackage[utf8]{inputenc}
\usepackage[T1]{fontenc}
\usepackage{float}

\usepackage{pifont}
\newcommand{\cmark}{\ding{51}}%
\newcommand{\xmark}{\ding{55}}%

\usepackage{xspace}
\makeatletter
\DeclareRobustCommand\onedot{\futurelet\@let@token\@onedot}
\def\@onedot{\ifx\@let@token.\else.\null\fi\xspace}

\def\eg{\emph{e.g}\onedot} 
\def\ie{\emph{i.e}\onedot} 
 
\def\etc{\emph{etc}\onedot}

\def\etal{\emph{et al}\onedot}
\def\paint{\emph{paint2pix} }   
\makeatother

\usepackage{multirow}
\usepackage[pagebackref,breaklinks,colorlinks]{hyperref}

\usepackage[width=122mm,left=12mm,paperwidth=146mm,height=193mm,top=12mm,paperheight=217mm]{geometry}

\begin{document}

\pagestyle{headings}
\mainmatter
\def\ECCVSubNumber{291}  

\title{Paint2Pix: Interactive Painting based Progressive Image Synthesis and Editing} %

\titlerunning{ECCV-22 submission ID \ECCVSubNumber} 
\authorrunning{ECCV-22 submission ID \ECCVSubNumber} 
\author{Anonymous ECCV submission}
\institute{Paper ID \ECCVSubNumber}

\titlerunning{Paint2Pix: Interactive Painting based Image Synthesis and Editing}

\author{Jaskirat Singh$^{1,2}$, Liang Zheng$^2$, Cameron Smith$^{1}$, Jose Echevarria$^1$\\
$^1$Adobe Research, $^2$Australian National University\\}

%
\authorrunning{J. Singh et al.}
%
\institute{}

\maketitle

\begin{abstract}

Controllable image synthesis with user scribbles is a topic of keen interest in the computer vision community. In this paper, for the first time we study the problem of photorealistic image synthesis from incomplete and primitive human paintings. In particular, we propose 
a novel approach 
\emph{paint2pix}, 
which learns to predict (and adapt) ``what a user wants to draw'' from rudimentary brushstroke inputs, by learning a mapping from the manifold of incomplete human paintings to their realistic renderings. When used in conjunction with recent works in autonomous painting agents, we show that \paint can be used for progressive image synthesis from scratch. During this process, \paint allows a novice user to progressively synthesize the desired image output, while requiring just few coarse user scribbles to accurately steer the trajectory of the synthesis process.
Furthermore, we find that our approach also forms a surprisingly convenient approach for real image editing, and allows the user to perform a diverse range of custom fine-grained edits through the addition of only a few well-placed 
brushstrokes. Source code and demo is available at \url{https://github.com/1jsingh/paint2pix}\footnote{Accepted to ECCV 2022. Supplemental video and demo available at \href{https://1jsingh.github.io/paint2pix}{project website}.}.

\end{abstract}

\section{Introduction}
\label{sec:intro}

\begin{figure}[t]
\begin{center}
\subfigure[Progressive image synthesis]{\includegraphics[width=0.65\linewidth]{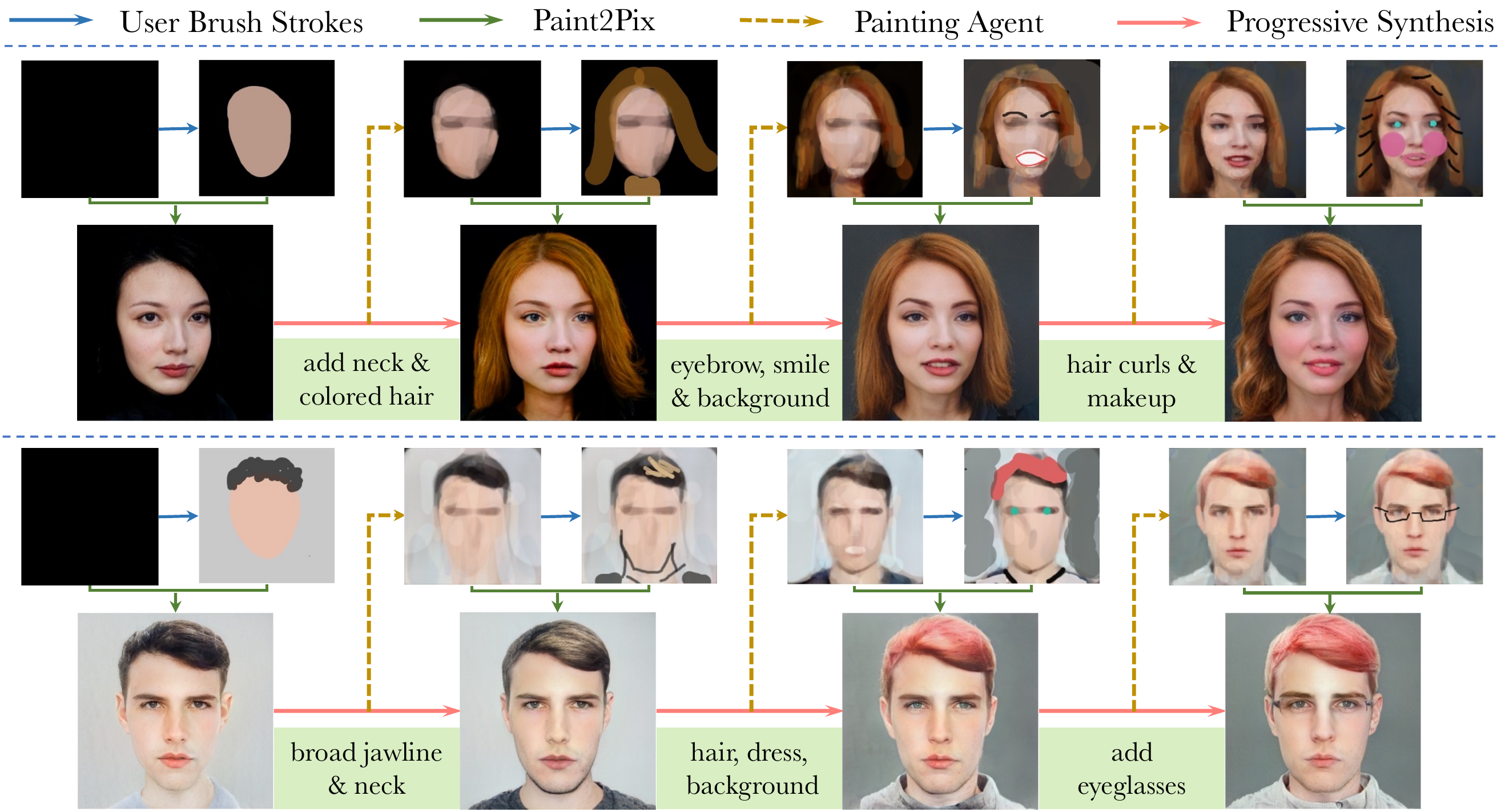}} \hfill
\subfigure[Real image editing]{\includegraphics[width=0.33\linewidth]{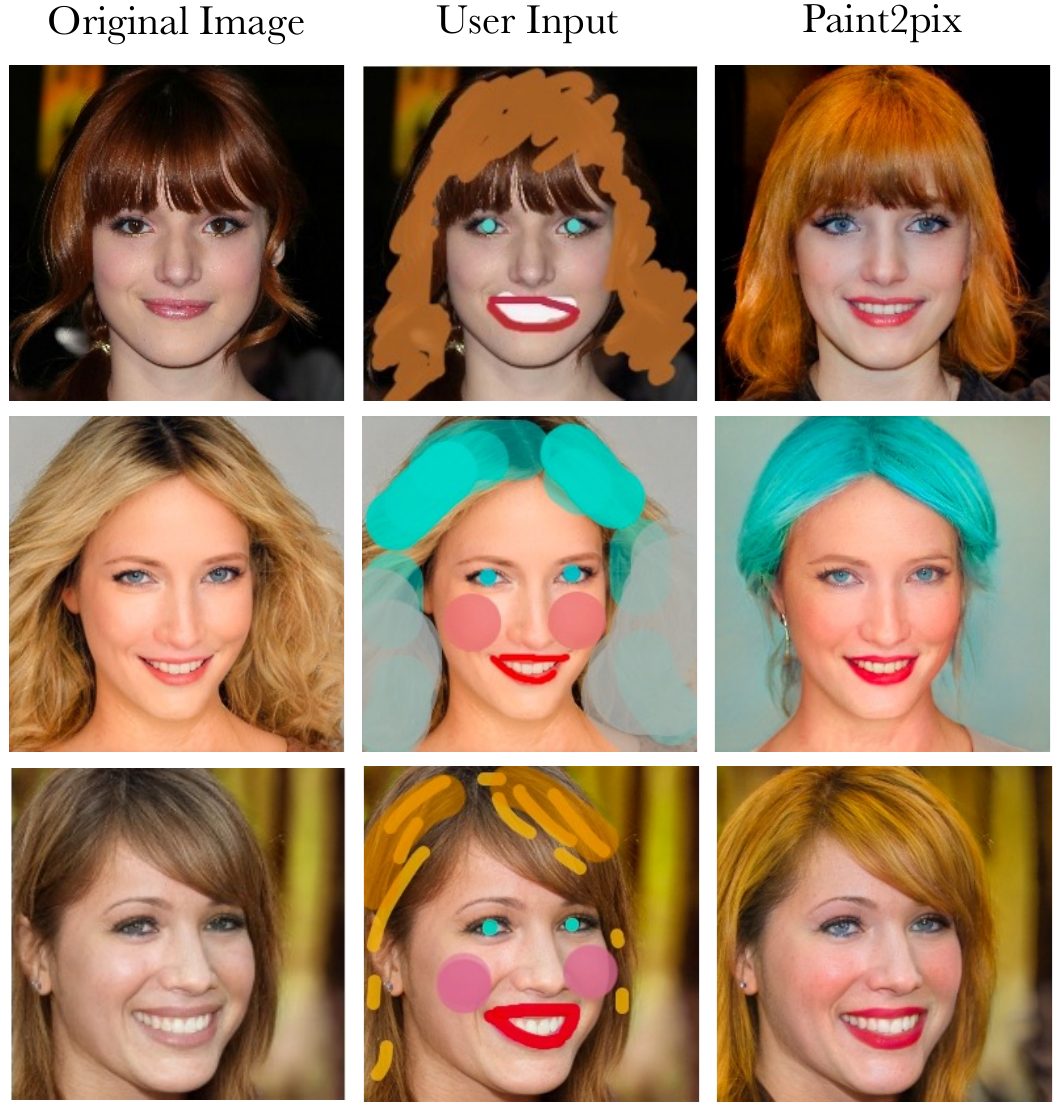}}
\vskip -0.1in
\caption{Overview. We propose \emph{paint2pix} which helps the user directly express his/her ideas in visual form by learning to predict user-intention from a few rudimentary brushstrokes. The proposed approach can be used for (a) synthesizing a desired image output directly from scratch 
wherein it allows the user to control the overall synthesis trajectory using just few coarse brushstrokes (blue arrows) at key points,
or, (b) performing a diverse range of custom edits directly on real image inputs.}
\label{fig:overview}
\end{center}
\vskip -0.3in
\end{figure}

The human painting process represents a powerful mechanism for the expression of our inner visualizations. However, accurate depiction of the same is often quite time consuming and limited to those with sufficient artistic skill. Conditional image synthesis provides a popular solution to this problem, and simplifies output image synthesis based on higher-level input modalities (segmentation, sketch) which can be easily expressed using coarse user scribbles. For instance, segmentation based image generation methods \cite{park2019semantic,zhu2020sean, choi2018stargan,choi2020stargan} allow for control over output image attributes based on user-editable semantic segmentation maps. However, they have obvious disadvantage of requiring large-scale dense semantic segmentation annotations for training, which makes them not easily scalable to new domains. Unsupervised sketch based image synthesis has also been explored \cite{chen2018sketchygan,liu2020unsupervised,ghosh2019interactive}, but they do not provide control over non-edge image areas.

In this paper, we explore the use of another modality in this direction, by studying the problem of photorealistic image synthesis from \emph{incomplete and primitive human paintings}. This is motivated from the observation that when constrained to a particular domain (\eg,  faces), a lot of information about the final image output can be inferred from fairly rudimentary and partially drawn human paintings. We thus propose a novel approach \emph{paint2pix}, which learns to predict (and adapt) ``what the user intends to draw'' from rudimentary brushstroke inputs, by learning a mapping from the manifold of incomplete human paintings to their realistic renderings. However, learning the
manifold of incomplete human paintings is challenging as it would require extensive collection of human painting trajectories for each target domain. 
For this challenge, we show that a fair approximation of this manifold can still be obtained by using painting trajectories from recent works on autonomous human-like painting agents \cite{singh2021intelli}.

While predicting photo-realistic outputs from partially drawn paintings might be helpful for capturing certain parts of a user's visualization (\eg,  face shape, hairstyle), fine grain control over different image attributes might be missing. In order to address this need for fine-grained control, we introduce an interactive synthesis strategy, wherein \paint when used in conjunction with an autonomous painting agent, allows a novice user to progressively synthesize and refine the desired image output using just few rudimentary brushstrokes. The overall image synthesis (refer Fig.~\ref{fig:overview}a) is performed in a progressive fashion wherein \paint and the autonomous painting agent are used in successive steps. Starting with an empty canvas, the user begins by making few rudimentary strokes (\eg, describing face shape, color) to obtain an initial user-intention prediction (through \emph{paint2pix}). The painting agent then uses this prediction to paint until a user-controlled timestep, at which point, the user again provides a coarse brushstroke input (\eg, describing finer details like hair color) to change the trajectory of the synthesis process. By iterating between these steps till the end of painting trajectory, the human artist is able to gain significant control over final image contents whilst requiring to input only few coarse scribbles (blue arrows in Fig.~\ref{fig:overview}a) at key points of the autonomous painting process.

In addition to progressive image synthesis, the proposed approach can also be used to perform fine-grained editing on real-images (Fig.~\ref{fig:overview}b). As compared with previous latent space manipulation methods \cite{patashnik2021styleclip,shen2021closed,abdal2021styleflow}, we find that our approach forms a surprisingly convenient 
alternative for making a diverse range of custom fine-grained modifications through the use of a few user scribbles. 
Furthermore, we show that custom edits (\eg, adding smile, changing makeup) are not limited to the image on which the modifications were performed but show generalization across the input domain (Sec.~\ref{sec:global_edits}). Put another way, once the user is satisfied with a custom edit on one image, the same edit can then be transferred to another image (from the same input domain) in a semantically-consistent manner.

To summarize, the main contributions of this paper are \textbf{1)} We introduce a novel task of photorealistic image synthesis from incomplete and primitive human paintings. \textbf{2)} We propose \paint which 
learns to predict (and adapt) 
``what a user wants to ultimately draw'' 
from rudimentary brushstroke inputs. \textbf{3)} We finally demonstrate the efficacy of our approach for (a) progressively synthesizing an output image from scratch, and, (b) performing a diverse range of custom edits directly on real image inputs.

\section{Related Work}
\label{sec:related_work}

\textbf{Autonomous painting agents.}
In recent years, substantial research efforts \cite{singh2021intelli,singh2021combining,huang2019learning,liu2021paint,zou2021stylized,wang2021self,kotovenko2021rethinking} have been focused on developing autonoumous painting agents which can learn an unsupervised stroke decomposition for the recreation of a given target image. 
Despite their efficacy, previous works in this area are often limited to the non-photorealistic recreation of \emph{a provided target image}. This assumes that the user already has a fixed reference image that he/she wants to recreate. However, in practical applications the intended image output may not be available and has to be synthesized in a progressive fashion. Our work thus proposes to develop a new application for autonomous painting agents by predicting user-intention from incomplete canvas frames.
\vskip 0.05in
\noindent \textbf{Segmentation based image generation.}
Image to image translation frameworks have been extensively studied for controllable generation of highly realistic image outputs based on a more simplified image representation. For instance, \cite{park2019semantic,zhu2020sean,lee2020maskgan, esser2021taming, isola2017image,liu2019learning, sushko2020you} use conditional generative adversarial networks for controllable image synthesis using user-provided semantic segmentation maps. While effective, these works require large-scale semantic segmentation annotations for training, which limits their scalability to new domains. Furthermore, making fine-grained changes within each semantic contour after image synthesis is non-trivial and often relies on style encoding methods \cite{zhu2020sean, choi2020stargan}, 
which require the user to first find a set of reference images which best describe the nature of each intended change (\emph{e.g.} adding makeup or changing hair style for facial images). In contrast, our work allows for a range of custom fine-grained image editions through the addition of just few well placed brush strokes.

\vskip 0.05in
\noindent \textbf{Sketch based image generation} 
has also been explored \cite{ghosh2019interactive,liu2020unsupervised,lee2020reference,li2020deep,chen2009sketch2photo,chen2018sketchygan,xiang2022adversarial,yang2021controllable}.
Ghosh \etal \cite{ghosh2019interactive} predict possible image outputs from rudimentary sketches of simple objects. Richardson \etal \cite{richardson2021encoding} use an encoder-decoder network to map input sketches to output images in a given domain. 
While effective in controlling initial aspects of the image output, the use of sketches (compared to paintings) for image generation is less effective as it offers limited control and sensitivity to changes made in non-edge areas. 

\vskip 0.05in
\noindent \textbf{GAN inversion.}
Interactive image generation and editing with user scribbles has also been explored 
in the context 
of GAN-inversion methods 
\cite{zhu2016generative,abdal2019image2stylegan,abdal2020image2stylegan++}. Zhu \etal \cite{zhu2016generative} propose a hybrid optimization approach for 
projecting user-given strokes onto the natural image manifold. Similarly, \cite{abdal2019image2stylegan,abdal2020image2stylegan++} use GAN-inversion to perform local image edits with user scribbles. While effective for small-scale photorealistic manipulations, these methods often lack means to learn the distribution of user-inputs (manifold of rudimentary paintings in our case) and thus are limited to performing a pure color-based optimization. As shown in Sec.~\ref{sec:comparisons}, this leads to poor performance on from-scratch synthesis and semantic edits on real images.

\begingroup
\renewcommand{\arraystretch}{1.1}
\setlength{\tabcolsep}{4.1pt}
\begin{table}[t]
\begin{center}
\small
\begin{tabular}{l|cccc}
\toprule
\emph{Method Attribute}
& \emph{Paint2Pix} & \emph{GAN-Inversion} & \emph{Seg2Photo} & \emph{Sketch2Photo}\\
\hline
\emph{From scratch} & \cmark & \xmark & \cmark & \cmark \\
\emph{Responsiveness} & \cmark & \cmark & \xmark & \xmark \\
\emph{No user-expertise} & \cmark & \xmark & \cmark & \cmark \\
\emph{Data efficiency}  & \cmark & \cmark & \xmark & \cmark \\
\bottomrule
\end{tabular}
\end{center}
\caption{Related work overview. Broad positioning of our work with respect to other methods for controllable image synthesis with user scribbles. (refer Sec.~\ref{sec:related_work} for details)}
\label{tab:perfomance_matrix}
\vskip -0.22in
\end{table}
\setlength{\tabcolsep}{5.1pt}
\endgroup

\vskip 0.05in
\noindent
\textbf{Positioning our work.} Table~\ref{tab:perfomance_matrix} summarizes the positioning of our approach with respect to previous methods performing controllable image synthesis using user-given brushstrokes/scribbles. In particular, we posit the comparative benefits of our approach with respect to the following desirable properties.
\begingroup
\begin{itemize}
    \item \emph{\textbf{Image synthesis from scratch}}. While Paint2pix, segmentation and sketch based methods allow for direct synthesis of the primary image from scratch, GAN-inversion methods perform a more color-based optimization and thereby show poor performance on image synthesis from scratch (Sec.~\ref{sec:from_scratch}).
    \item \emph{\textbf{Responsiveness (control) over all image areas}}. Due to the one-to-many nature of learned mappings, segmentation based methods fail to provide fine-grained control over attributes within each semantic region. Similarly, sketch-based methods lack sensitivity to changes in non-edge areas.
    \item \emph{\textbf{Usability by novice artists}}. A key advantage of our method is that it allows a novice artist to control the synthesis process while using fairly rudimentary brushstrokes. In contrast, GAN-inversion based methods require the user to make sufficiently detailed strokes in order to preserve closeness to the real image manifold (refer Sec.~\ref{sec:fine-grain} for more details).
    \item \emph{\textbf{Data efficiency}}. Our method is completely self-supervised and uses \cite{singh2021intelli} to approximate the manifold of incomplete human paintings. In contrast, segmentation based methods require large-scale dense semantic maps for training on each target domain, which limits their scalability.
\end{itemize}
\endgroup

\begin{figure}[t]
\vskip -0.1in
\begin{center}
\centerline{\includegraphics[width=0.95\linewidth]{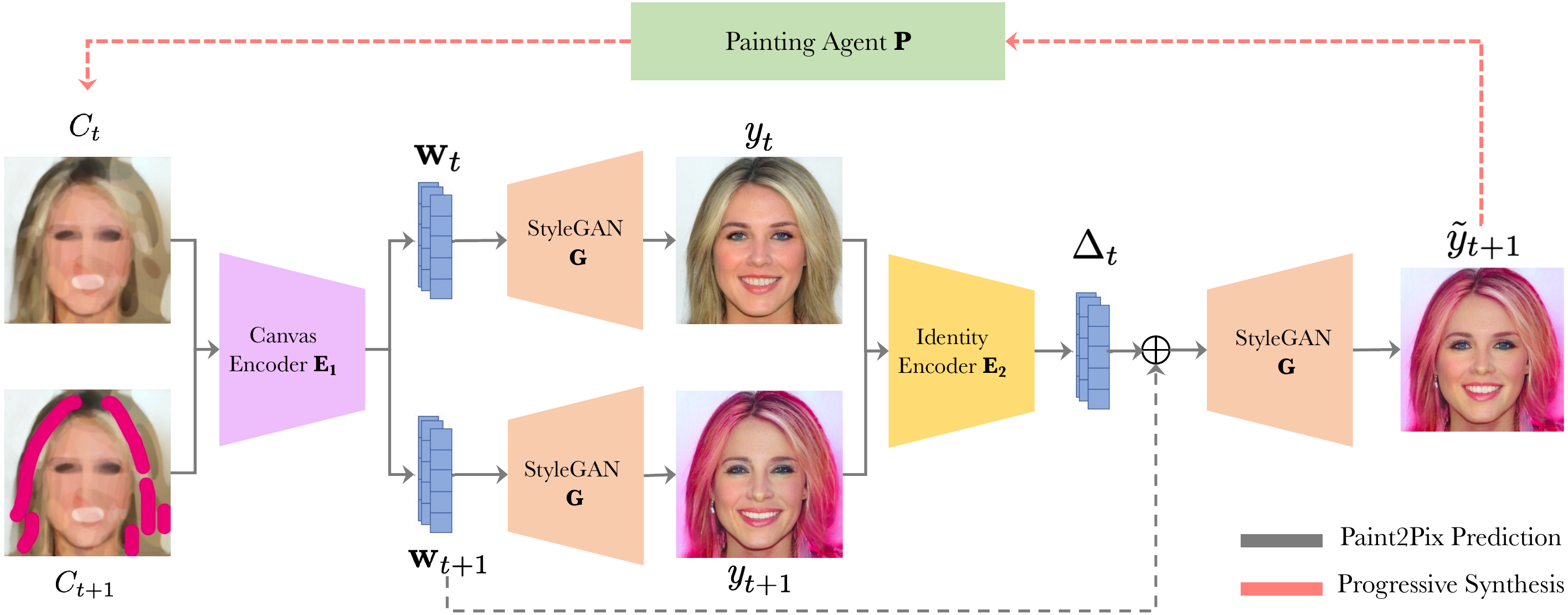}}
\caption{Model Overview. The \paint model helps simplify the image synthesis task by predicting user-intention from rudimentary canvas state $C_t$, while also allowing the user to accurately steer the synthesis trajectory using coarse brushstroke inputs in $C_{t+1}$. This is done in two steps. First, the canvas encoder $\mathbf{E_1}$ learns a mapping between the manifold of incomplete paintings and real images to predict realistic user-intention predictions $\{y_t,y_{t+1}\}$ from $\{C_t,C_{t+1}\}$ respectively. These intermediate predictions are then fed into a second identity encoder $\mathbf{E}_2$ to predict a latent-space correctional term $\Delta_t$, which ensures that the final prediction $\tilde{y}_{t+1}$ preserves the identity of the prediction from the original canvas $C_t$, while at the same time incorporating changes made by the user input brushstrokes in $C_{t+1}$. The progressive synthesis process can then be continued by feeding final prediction $\tilde{y}_{t+1}$ to an autonomous painting agent which paints it till a user-controlled timestep, at which point, the user can again add coarse brushstroke inputs in order to better express her inner ideas in the final image output.}
\label{fig:model_design}
\end{center}
\vskip -0.4in
\end{figure}

\section{Our Method}
\label{sec:method}

The \paint model uses a two step decoupled encoder-decoder architecture (refer Fig.~\ref{fig:model_design}) for predicting user intention from incomplete user paintings, while at the same time allowing a novice user to control the overall image synthesis trajectory using coarse and rudimentary brushstroke inputs.

\subsection{Canvas Encoding Stage}
\label{sec:canvas_encoding}

The goal of the canvas encoding stage is two-fold: 1) predict user-intention by learning a mapping between the manifold of incomplete user paintings to their realistic output renderings, while at the same time 2) allow for modification in the progressive synthesis trajectory based on coarse user-brushstrokes.

In particular, given current canvas state $C_t$ and the updated canvas state after coarse user-brushstroke input $C_{t+1}$, we first use a canvas encoder $\mathbf{E}_1$ to predict a tuple of initial latent vector predictions $\{\mathbf{w}_t, \mathbf{w}_{t+1}\}$ as,
\begin{align}
    \{\mathbf{w}_t, \mathbf{w}_{t+1}\} = \mathbf{E}_{1} (C_t,C_{t+1}).
\end{align}
These latent predictions are then fed into a StyleGAN \cite{karras2020analyzing} decoder network $\mathbf{G}$, in order to get realistic user-intention predictions $\{y_t,y_{t+1}\}$ corresponding to input canvas tuple $\{C_t,C_{t+1}\}$ respectively,
\begin{align}
    \{\mathbf{y}_t, \mathbf{y}_{t+1}\} = \{G(\mathbf{w}_t), G(\mathbf{w}_{t+1})\}.
\end{align}

\subsubsection{Losses.} Given realistic output ground-truth annotation $\hat{y}_t$ corresponding to canvas $C_t$ (refer Sec.~\ref{sec:overall_training}), the canvas encoder $\mathbf{E}_1$ is trained to learn to predict user-intention with the following prediction loss $\mathcal{L}_{pred}$,
\begin{align}
    \mathcal{L}_{pred} = \mathcal{L}_2(y_t,\hat{y}_t) + \lambda_1 \mathcal{L}_{lpips}(y_t,\hat{y}_t) + \lambda_2 \mathcal{L}_{id}(y_t,\hat{y}_t), 
\end{align}
where $\mathcal{L}_{lpips}$ is the perceptual similarity loss \cite{zhang2018perceptual} and $\mathcal{L}_{id}$ represents the Arcface \cite{deng2019arcface} / MoCo-v2 \cite{chen2020mocov2} features based identity similarity loss from Tov \etal \cite{tov2021designing}.


As previously mentioned, we would also like to ensure that the output predictions are modified in order to reflect the changes added by the user in $C_{t+1}$. This is then achieved by the following edition loss $\mathcal{L}_{edit}$,
\begin{align}
    \mathcal{L}_{edit} = \mathcal{L}_{lpips}(\Delta C_t ,\Delta y_t) + \lambda_3 \mathcal{L}_{adv}(w_{t+1}) +  \lambda_4 \Vert w_{t+1}-w_{t} \Vert_2, \label{eq:l_edit}
\end{align}
where $\Delta C_t = C_{t+1} - C_{t}$ and $\Delta y_t = y_{t+1} - y_{t}$ represent the changes in the original canvas and output predictions respectively. $\mathcal{L}_{adv}$ refers to the latent discriminator loss from e4e \cite{tov2021designing} to ensure realism of the latent space prediction. Finally, the last term ensures that the codes $\{w_{t},w_{t+1}\}$ for consecutive image outputs $\{y_{t},y_{t+1}\}$ lie close in the StyleGAN \cite{karras2020analyzing} latent space. 

The overall loss for the canvas encoding stage is then defined as follows,
\begin{align}
    \mathcal{L}_{canvas} = \mathcal{L}_{pred} + \lambda_{edit}\  \mathcal{L}_{edit}.
\end{align}

\subsection{Identity Embedding Stage}
\label{sec:id_embed}

While enforcing closeness of consecutive latent vector codes $\{w_{t},w_{t+1}\}$(Eq.~\ref{eq:l_edit}) , helps in ensuring that the updated output prediction $y_{t+1}$ is derived from the original prediction $y_t$, inconsistencies might still arise due to subtle changes in the identity of the underlying prediction (Fig.~\ref{fig:model_design}). 
Thus, the goal of the second stage is to preserve the underlying identity between consecutive image predictions and thereby ensure semantic consistency of the overall image synthesis process.


To address this, we train a second identity encoder $\mathbf{E}_2$ which ensures that the final prediction $\tilde{y}_{t+1}$ preserves identity of the original prediction $y_t$ while still reflecting the changes made by the user in canvas $C_{t+1}$. In particular, given output image predictions $\{y_t, y_{t+1}\}$ from the canvas encoding stage, the identity encoder $\mathbf{E}_2$ predicts a correctional term $\Delta_t$ to update the latent codes as,
\begin{align}
    \tilde{w}_{t+1} = w_{t+1} + \Delta_t, \quad \mbox{where} \quad \Delta_t = \mathbf{E}_2(y_t, y_{t+1}).
\end{align}
The updated latent code $\tilde{w}_{t+1}$ is then used to predict the final output prediction $\tilde{y}_{t+1}$ using the StyleGAN \cite{karras2020analyzing} decoder $\mathbf{G}$ as,
\begin{align}
    \tilde{y}_{t+1} = \mathbf{G}(\tilde{w}_{t+1}).
\end{align}

\subsubsection{Losses.} The identity encoder is trained using the following loss, 
\begin{align}
    \mathcal{L}_{embed} = \mathcal{L}_{2}(y_{t+1},\tilde{y}_{t+1}) + \lambda_5 \mathcal{L}_{lpips}(y_{t+1},\tilde{y}_{t+1})  + \lambda_6 \Vert \Delta_t \Vert_2 + \lambda_7 \mathcal{L}_{id}(y_{t},\tilde{y}_{t+1})
\end{align}
where the first three terms ensure the preservation of edits made by the user in $C_{t+1}$, while the last term enforces that the final prediction $\tilde{y}_{t+1}$ preserves the identity of the original image prediction $y_{t}$, thereby ensuring consistency of the overall progressive synthesis process.

\subsubsection{Reason for decoupled encoders.} 
While its feasible to design a model architecture wherein both $\mathcal{L}_{canvas}$ and $\mathcal{L}_{embed}$ are applied using a single encoder, the use of a decoupled identity encoder offers several practical advantages. For instance, while ensuring identity consistency is usually important (\eg,  making fine-grained changes), a change in underlying identity might sometimes be actually desirable, especially at the beginning of the progressive synthesis process. The decoupling of canvas encoding and identity embedding stage is therefore useful, as it allows the user to apply identity correction depending on the nature of the intended change. Furthermore, as shown in Sec.~\ref{sec:multi-modal}, decoupling the two stages allows our model to perform multi-modal synthesis without requiring any special architecture for producing multiple output predictions.

\subsection{Overall Training}
\label{sec:overall_training}

\subsubsection{Total loss.} The overall \paint model is jointly trained using both canvas-encoding $\mathcal{L}_{canvas}$ and identity embedding $\mathcal{L}_{embed}$ losses:
\begin{align}
    \mathcal{L}_{total} = \mathcal{L}_{canvas} + \lambda_{embed} \  \mathcal{L}_{embed}.
\end{align}

\subsubsection{Ground truth painting annotations.}
\label{sec:painting_annotations}

As discussed before, a key requirement of our approach is the ability to learn a mapping between the manifold of incomplete human-user paintings to their ideal realistic outputs. This requirement is challenging as it would need large-scale collection of human painting trajectories for each target domain, making our method intractable for most practical applications. To address this, we propose to instead use the recent works on autonomous painting agents for obtaining a decent approximation for the manifold of incomplete user paintings. The accuracy of such an approximation would depend highly on the domain gap between the incomplete paintings made by human users as compared to those made by a painting agent. We reduce this domain gap by using the recently proposed \emph{Intelli-paint} \cite{singh2021intelli} method, which has been shown to generate intermediate canvas frames which are more intelligible to actual human artists as opposed to previous works \cite{liu2021paint,zou2021stylized,singh2021combining,huang2019learning}.

In particular, for each painting trajectory trying to recreate a given target image $I_{target} \in \mathcal{D}$ ($\mathcal{D}$ is input domain, \eg,  FFHQ \cite{karras2019style} for faces), we collect input canvas annotations by uniformly sampling 20 tuples of consecutive canvas frames $\{C_t,C_{t+1}\}$ observed during the painting process. The output image annotation $\hat{y}_t$ for all sampled canvas tuples (from the same trajectory) is then set to the original target image $I_{target}$. Furthermore, we collect painting annotations under various brushstroke counts $N_{strokes} \in [200, 500]$, as it helps capture the diverse degrees of abstraction observed in paintings made by actual human artists.

\section{Paint2pix for Progressive Image Synthesis}

In this section, we demonstrate how the one-step user intention prediction network from Sec.~\ref{sec:method} can be used for progressively synthesizing a potential user's ideas as a realistic image output without requiring artist-level expertise.

\begin{figure}[h!]
\vskip -0.0in
\begin{center}

\subfigure[Progressive Image Synthesis for Cars (Stanford-Cars \cite{KrauseStarkDengFei-Fei_3DRR2013}) domain. ]{\includegraphics[width=0.9\linewidth]{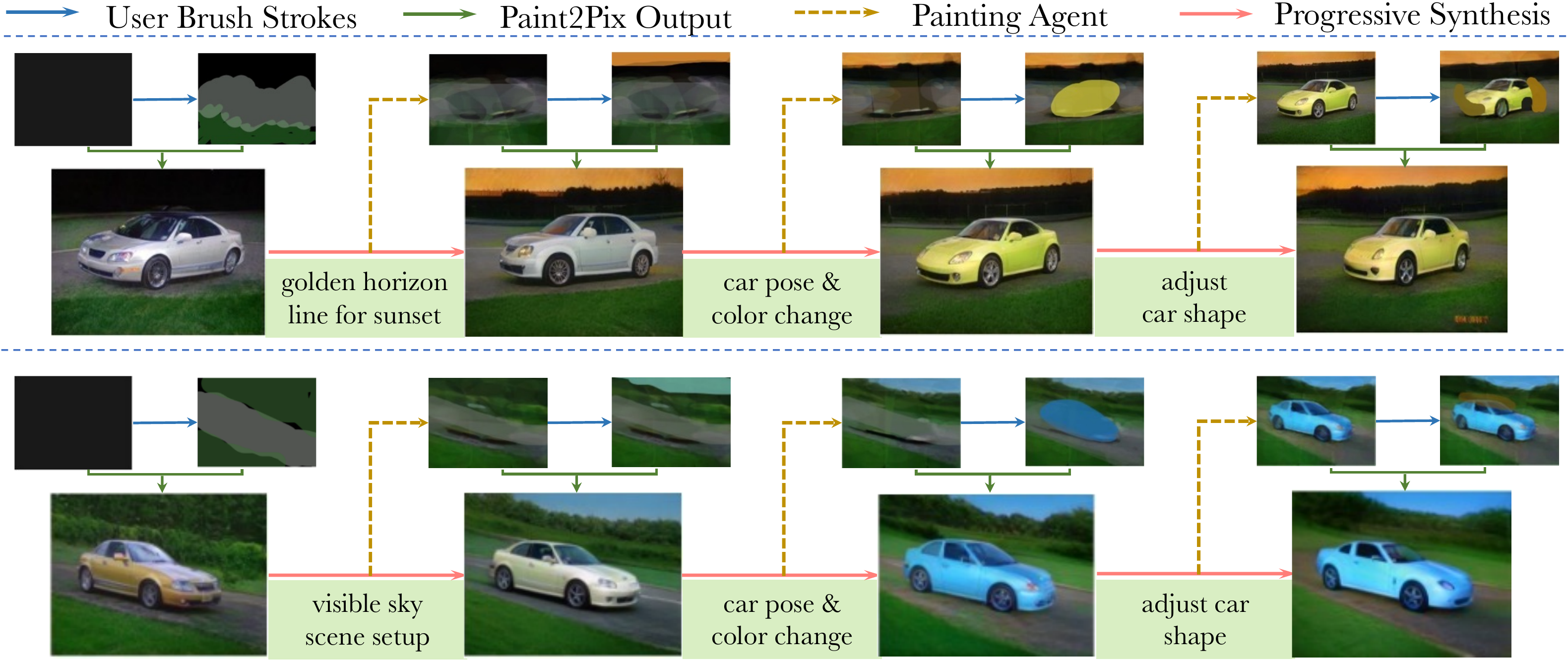}} 
\subfigure[Progressive Image Synthesis for Facial (FFHQ \cite{karras2019style}) domain.]{\includegraphics[width=0.9\linewidth]{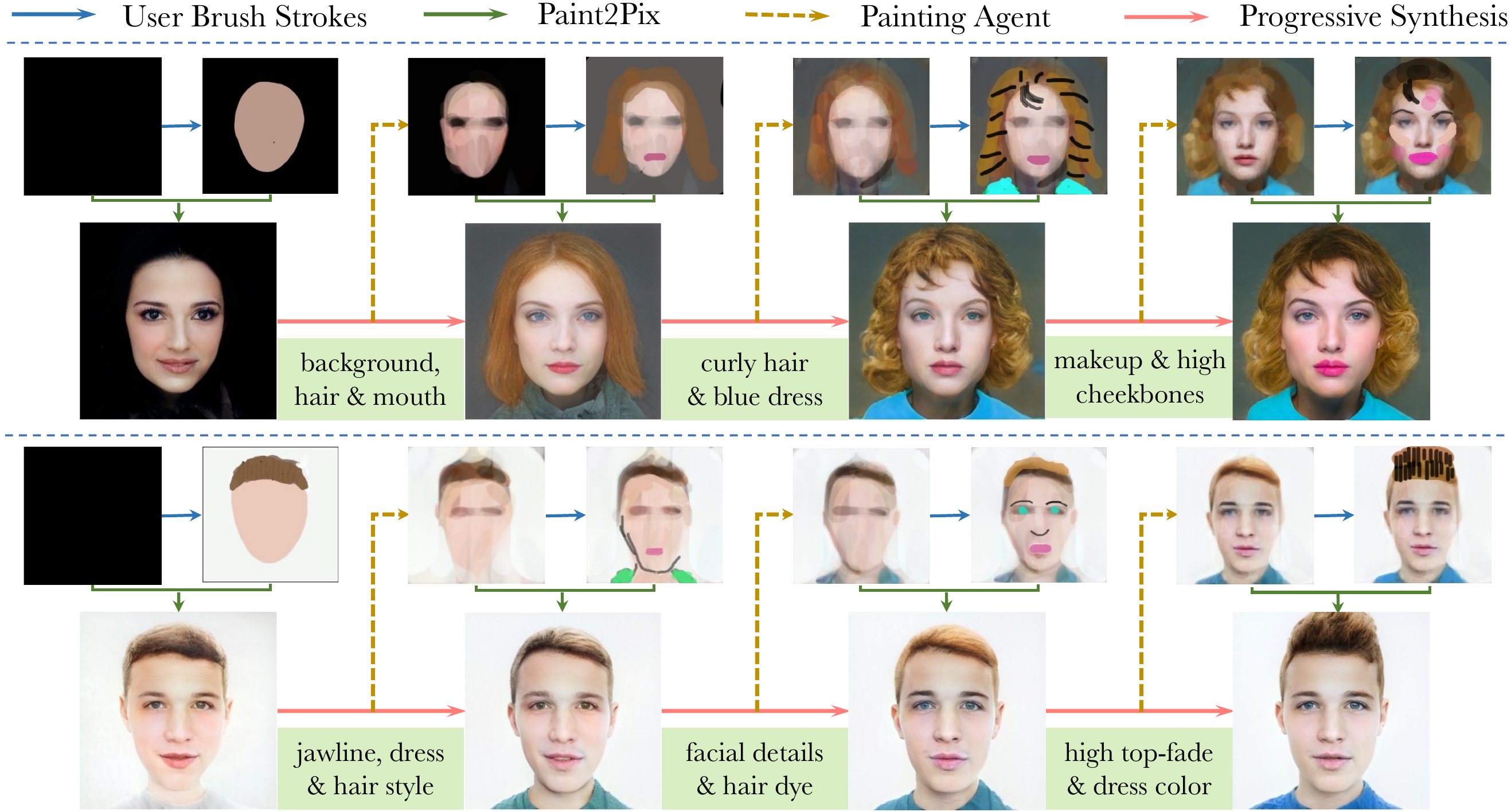}} 
\vskip -0.05in
\caption{Paint2pix for Progressive Image Synthesis.}
\label{fig:prog_synthesis}
\end{center}
\vskip -0.4in
\end{figure}

Figure~\ref{fig:prog_synthesis} demonstrates the use of \paint for progressive image synthesis from scratch. A potential user would start the painting process by adding a few rudimentary brushstrokes on the canvas (\eg, background scene for cars or face shape, color for faces). The \paint network then outputs a set of possible realistic image renderings (refer Sec.~\ref{sec:multi-modal} for more details on multi-modal synthesis) that the user might be interested in drawing. The user may then select the image that most closely resembles his/her idea to obtain a user-intention prediction. The progressive synthesis process can then be continued by feeding this prediction to an autonomous painting agent which paints it till a user-controlled timestep, at which point, the user can again add coarse scribbles (\eg,  describing finer details like sky color for cars or hairstyle for faces) in order to steer the synthesis trajectory according to his/her ideas. By continuing this iterative process till the end of the painting process, a novice user can gain significant control over the final image contents while requiring to only input few rudimentary brushstrokes at key points in the autonomous painting trajectory.

\begin{figure}[t]
\begin{center}
\centerline{\includegraphics[width=0.9\linewidth]{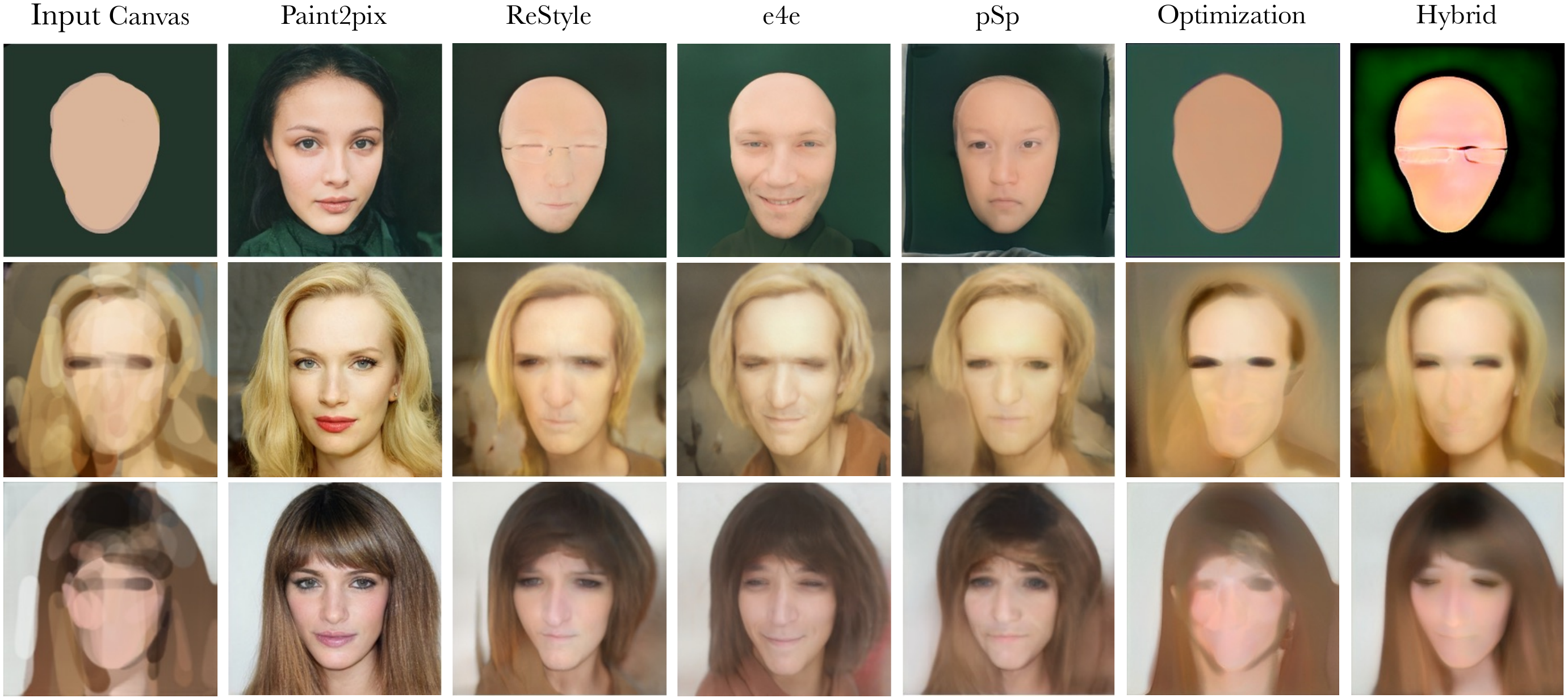}}
\vskip -0.1in
\caption{Qualitative comparison: Predicting user intention from rudimentary paintings.}
\label{fig:from_scratch}
\end{center}
\vskip -0.4in
\end{figure}

\section{Comparison with Inversion Methods}
\label{sec:comparisons}

Interactive image generation and editing with user brushstrokes has also been explored in the context of GAN-inversion methods \cite{zhu2016generative,abdal2019image2stylegan,abdal2020image2stylegan++}, which use an encoder or optimization based inversion approach in order to project user scribbles onto the real image manifold.
In this section, we present extensive quantitative and qualitative results comparing our approach with existing GAN-inversion methods for image manipulation with user-scribbles. In particular, we demonstrate the efficacy of our approach in terms of both 1) from scratch synthesis: \ie, predicting user intention from fairly rudimentary paintings (Sec.~\ref{sec:from_scratch}), and 2) real image editing: allowing a potential user to make a range of custom fine-grain edits directly by just using a few coarse input brushstrokes (Sec. \ref{sec:fine-grain}). 

\vskip 0.05in
\noindent \textbf{Baselines.} We compare our results with recent state-of-the-art encoder based methods from Restyle \cite{alaluf2021restyle}, e4e \cite{tov2021designing}  and pSp \cite{richardson2021encoding}. In addition, we report results for \emph{optimization} based encoding approach from Karras \etal \cite{karras2020analyzing} and \emph{hybrid} strategy from Zhu \etal \cite{zhu2016generative}. Please note that in order to get best output quality, results for \cite{zhu2016generative} are reported while using a pretrained ReStyle \cite{alaluf2021restyle} encoder.

\subsection{Predicting User-Intention from Rudimentary Paintings}
\label{sec:from_scratch}

\vskip 0.05in
\noindent \textbf{Qualitative results.} Fig.~\ref{fig:from_scratch} shows qualitative comparisons while predicting photo-realistic (user-intention) outputs from rudimentary paintings. We clearly see that our approach results in much more photorealistic predictions for the user-intended final output. In contrast, the predominantly color-based optimization nature of previous GAN-inversion works leads to non-photorealistic projections when all color details are yet to be added by the user. For instance, while drawing a human face, it is quite common for an artist to first draw a coarse brushstroke for the face region without adding in the finer facial details. However, this leads to poor performance while using color-based optimization as it leads the model to instead predict an output face where the finer facial details are hardly noticeable. Adversarial loss in e4e \cite{richardson2021encoding} helps improve the realism of output images but it still performs worse than \paint for this task.

\vskip 0.05in
\noindent \textbf{Quantitative results.} We also report quantitative results for this task (and image editing tasks from Sec.~\ref{sec:fine-grain}) in Table \ref{tab:quant_results}. Results are reported in terms of the Fr\'echet inception distance (FID) \cite{heusel2017gans}, which is used capture the output image quality from different methods. Furthermore, we perform a human user-study (details in supp. material) and report the percentage of human users which prefer our method as opposed to competing works. As shown in Table \ref{tab:quant_results}, we observe that \paint produces better quality images (lower FID scores) and is preferred by majority of human users over competing methods.

\subsection{Real-image Editing}
\label{sec:fine-grain}

In addition to being able to perform progressive synthesis from scratch, \paint also offers a surprisingly convenient approach for making a diverse range of custom semantic edits (\eg,  add smile for faces) on real images by simply initializing the canvas input $C_t$ with a real image. We next compare our method with previous GAN-inversion works on performing real image editing with user scribbles.

\begin{figure}[h!]
\vskip -0.2in
\begin{center}
\centerline{\includegraphics[width=0.9\linewidth]{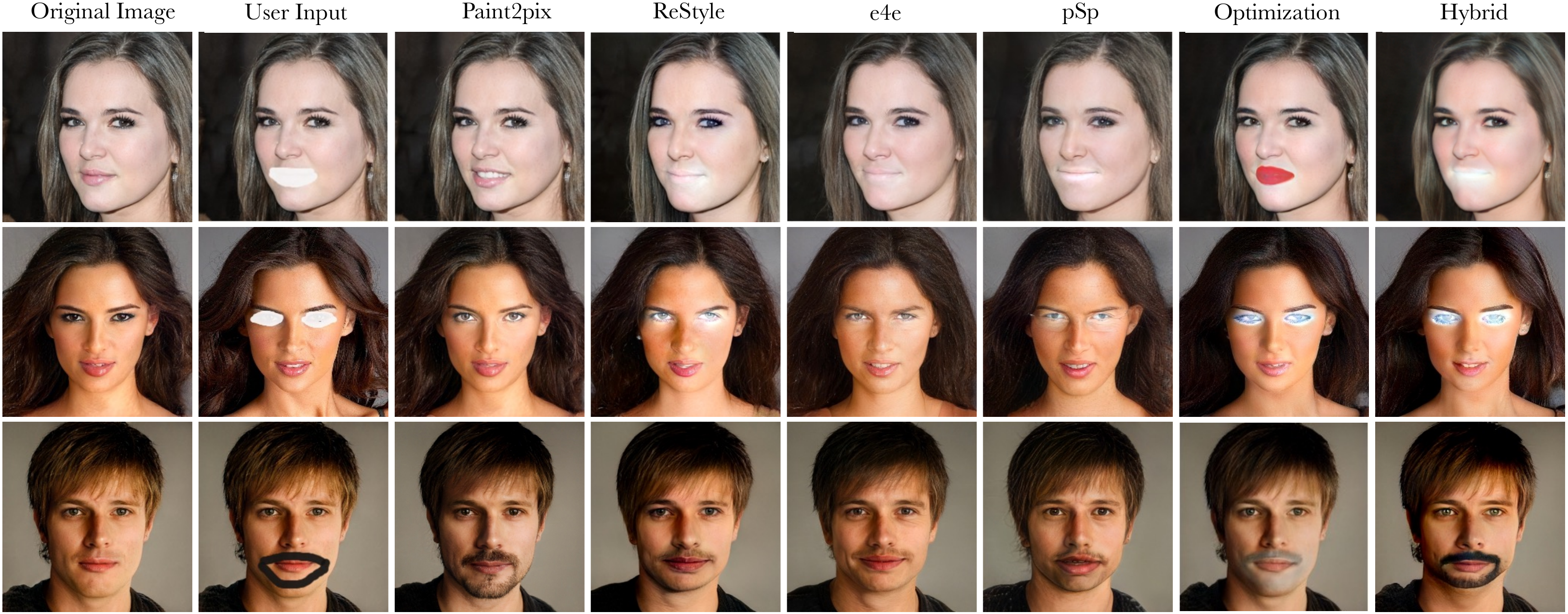}}
\vskip -0.1in
\caption{Paint2pix for achieving semantic image edits. Best viewed zoomed-in.}
\label{fig:semantic_edits}
\end{center}
\vskip -0.5in
\end{figure}

\vskip 0.05in
\noindent \textbf{Semantic image edits.} As shown in Fig.~\ref{fig:semantic_edits}, we observe that our approach performs much better when the nature of the underlying edit is not purely color-based. For instance, consider the the first example from Fig.~\ref{fig:semantic_edits}. Our method is able to correctly interpret that coarse white brushstrokes near the mouth region implies that the user is trying to add smile to the underlying facial image. In contrast, due to the predominantly color-based-optimization, gan-inversion methods fail to understand the change in semantics of face, and thus predict output faces in which the mouth region has been artificially-colored white.

\vskip 0.05in
\noindent \textbf{Color-based custom edits.} Even when the custom-edits are color-based, we show that \paint leads to outputs which are 1) more photorealistic, 2) exhibit a greater level of detail at the edit locations, 3) modify non-edit locations (in addition to edit locations) in order to maintain coherence of the resulting image and 4) better preserve the identity of the original image input.

\begin{figure}[h!]
\vskip -0.2in
\begin{center}
\centerline{\includegraphics[width=0.9\linewidth]{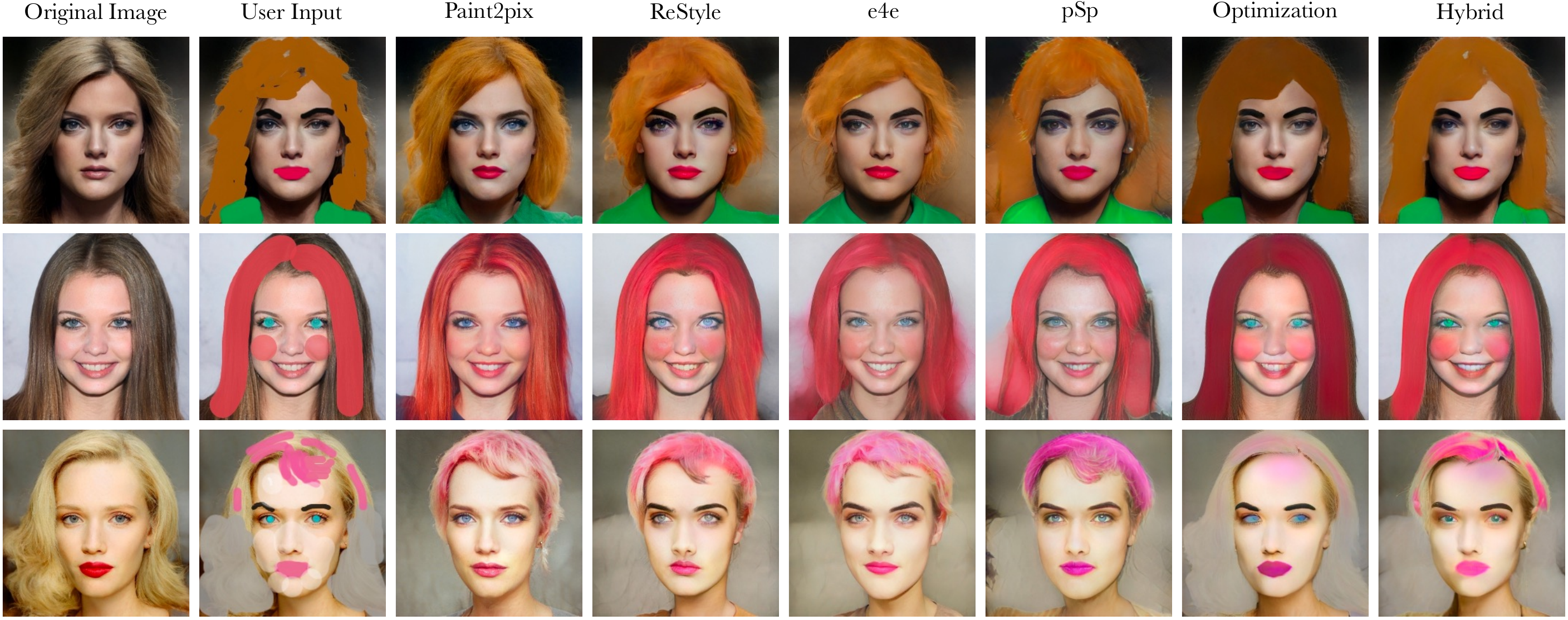}}
\vskip -0.1in
\caption{Paint2pix for achieving color-based custom edits. Best viewed zoomed-in.}
\label{fig:custom-color-edits}
\end{center}
\vskip -0.61in
\end{figure}

Results are shown in Fig.~\ref{fig:custom-color-edits}. Consider the first example (row-1). The increased realism of \paint outputs can be clearly seen by the more photorealistic and detailed representation at edit locations (\eg,  hair, eyebrows). Furthermore, note that our method shows a more global understanding of image semantics and subtly modifies the skin tone and the eye shading of the face in order to maintain consistency with user-given edits. In contrast, the color-based optimization of GAN-inversion methods exhibit a lower level of detail at edit locations (\eg,  hair in row 1-3 and makeup in row 2,3). Furthermore, we find that our method shows better performance in preserving the identity of the original input image in the final output (\eg,  row 2,3), which is highly essential for real-image editing.

\begingroup
\renewcommand{\arraystretch}{1.5}
\setlength{\tabcolsep}{3.8pt}
\begin{table}[t]
\vskip -0.05in
\begin{center}
\scriptsize
\begin{tabular}{l|cccccc|c}
\toprule
\multirow{2}{*}{Task} & \multicolumn{6}{c|}{FID ($\downarrow$) comparison} & User Study\\
\cline{2-8}
& Paint2pix & Restyle & e4e & pSp & Optimization & Hybrid & Paint2pix Preference\\
\hline
From scratch & \textbf{40.96} & 85.98 & 79.69 & 89.83 & 107.2 & 91.62 & 97.32 \%\\
Semantic Edits & \textbf{40.24} & 45.27 & 42.32 & 46.08 & 47.16 & 49.29 & 94.04\%\\
Color Edits & \textbf{63.56} & 100.2 & 93.11 & 107.3 & 116.4 & 114.2 & 93.85\%\\
\bottomrule
\end{tabular}
\end{center}
\vskip -0.07in
\caption{Quantitative evaluation. Col 2-7: FID results for comparing output image quality on different image synthesis, editing tasks. Col-8: Human user-study results reporting percentage of users which prefer Paint2pix outputs over other methods.}
\label{tab:quant_results}
\vskip -0.32in
\end{table}
\endgroup

\section{Inferring Global Edit Directions}
\label{sec:global_edits}
We next show that the custom edits (\eg, adding glasses, changing makeup) learned through \paint are not limited to the image on which the modifications were originally performed but instead show semantically-consistent generalization across the input domain. Put another way, once the user is satisfied with the output of a given custom edit on one image, the same edit can then be applied across different images from the input data distribution without requiring the user to repeat similar brushstrokes on each individual image.

In particular, consider $\{x_0,x_1\}$ be the original and edited image tuple with stylegan latent space vectors $\{\mathbf{w}_0,\mathbf{w}_1\}$ respectively. The custom edit $x_0 \rightarrow x_1$ can then be applied to another image $x$ (with stylegan latent code $\mathbf{w}$) by computing a modified latent space edit direction $\delta_{edit}(x)$ as,
\begin{align}
    \delta_{edit}(x) = \delta_{edit}(x_0) + \mathbf{E}_2(x,\mathbf{G}(\mathbf{w} + \delta_{edit}(x_0))), 
\end{align}
where $\delta_{edit}(x_0) = \mathbf{w}_1 - \mathbf{w}_0$ represents the original edit direction from $x_0 \rightarrow x_1$, and the second term ensures identity preservation in the transferred edit.

The original edit can then be transferred to the input image $x$ as,
\begin{align}
    x' = \mathbf{G}(\mathbf{w} + \alpha \ \delta_{edit}(x)),
\end{align}
where $\alpha$ is the edit strength and $\mathbf{G}$ is the styleGAN \cite{karras2020analyzing} decoder network. 

\begin{figure}[t]
\begin{center}
\centerline{\includegraphics[width=0.85\linewidth]{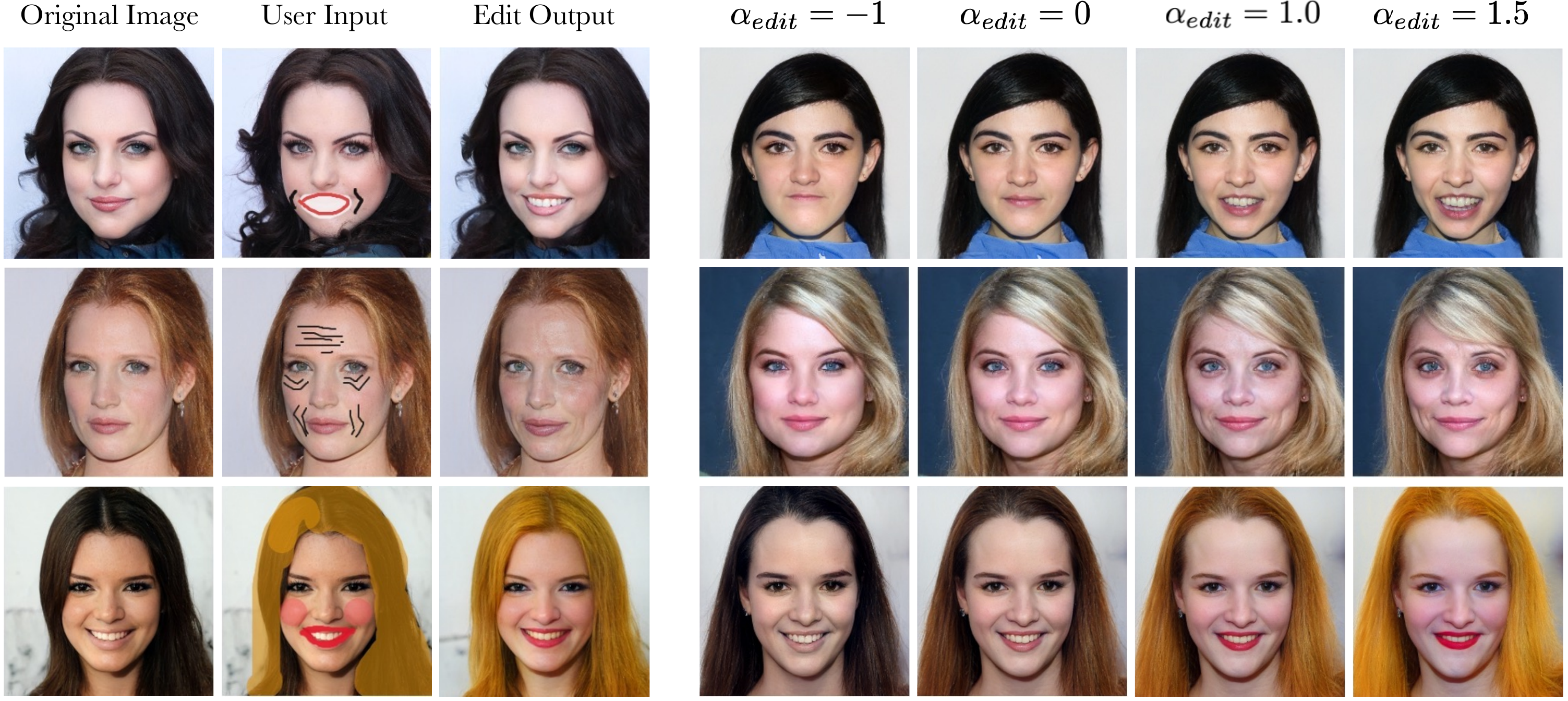}}
\vskip -0.1in
\caption{Inferring global edit directions using Paint2pix. Left: Original custom edit using Paint2pix. Right: Same edit transferred to another image with different edit strengths.}
\label{fig:global_edits}
\end{center}
\vskip -0.4in
\end{figure}

Results are shown in Fig.~\ref{fig:global_edits}. We are able to clearly see that custom edits learned on one image can be easily extended to different images in a semantic-consistent manner. Our experiments also reveal that the generalizability of the edit is largely independent of the complexity of the underlying edit. This enables us to transfer both simple (\eg, smile in row-1) and complex edits (\eg, makeup change in row-3) to new images across the input domain.

Furthermore, we observe that the strength of the intended edit can be varied by simply adjusting the edit-strength parameter $\alpha$. This helps us to use extrapolation in order to achieve edits which would be otherwise difficult to draw using rudimentary brushstrokes. For instance, while adding smile (using white brushstrokes) is easy, drawing a fully laughing face might be difficult for a novice artist. However, the same can be easily achieved by using a higher edit strength which allows us to extrapolate the original smiling edit to a laughing face edit (refer row-1,  Fig.~\ref{fig:global_edits}). Similarly, different levels of facial wrinkles (or aging) can also be achieved in an analogous fashion (refer row-2,  Fig.~\ref{fig:global_edits}).

\section{Multi-modal Synthesis}
\label{sec:multi-modal}
Predicting a single output for inferring user intention from an incomplete painting might not be always useful if the user's ideas are vastly different from the output prediction. The use of decoupled encoders in \paint is helpful in this regard, as it allows our approach to perform multi-modal synthesis for the final output without requiring special architecture changes. 

\begin{figure}[t]
\vskip -0.1in
\begin{center}
\subfigure[]{\includegraphics[width=0.44\linewidth]{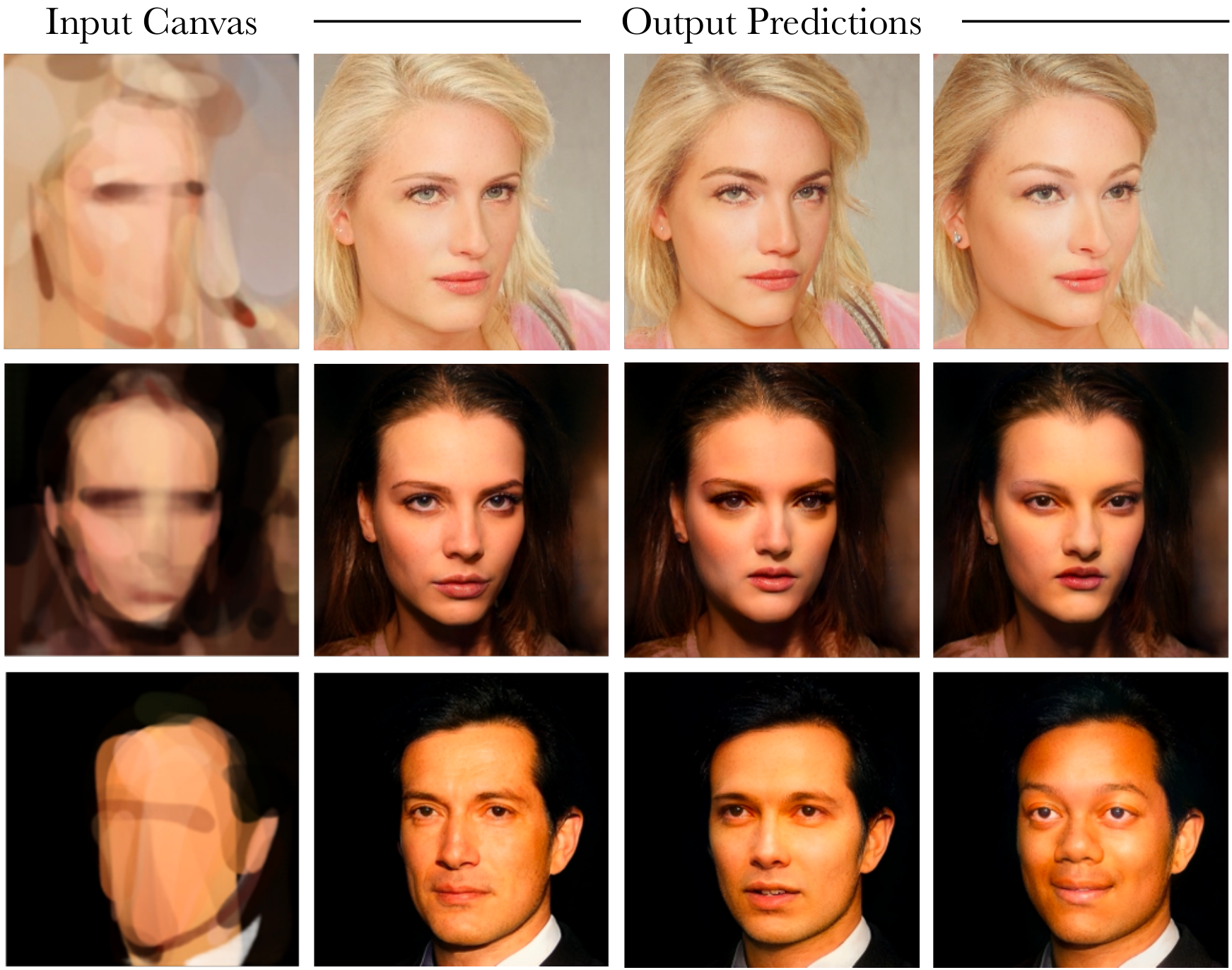}} 
\hfill
\subfigure[]{\includegraphics[width=0.54\linewidth]{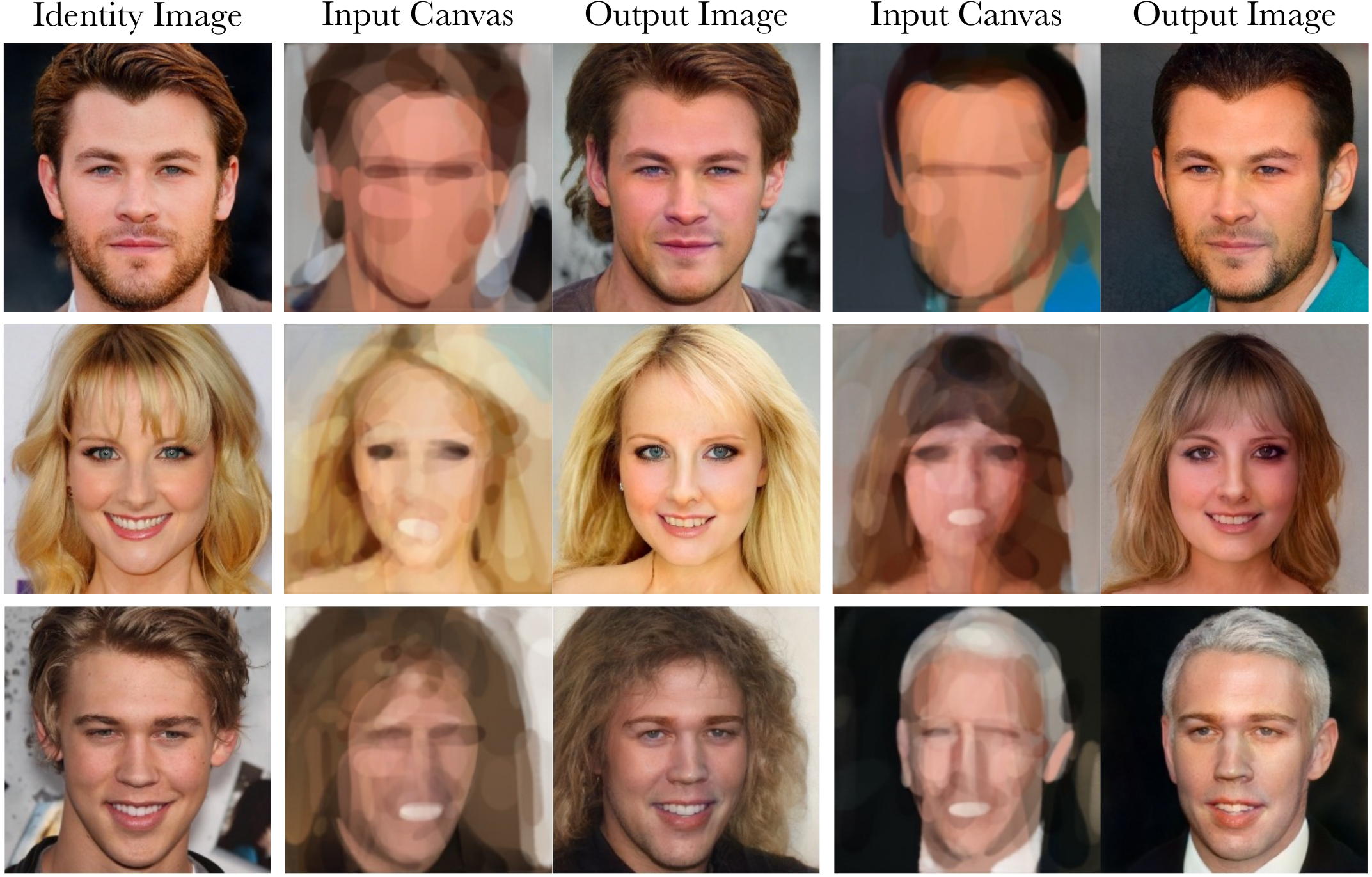}} 
\vskip -0.1in
\caption{Paint2pix for (a) multi-modal synthesis and (b) identity-conditioned generation.}
\label{fig:multi-modal}
\end{center}
\vskip -0.3in
\end{figure}

In practice, given an incomplete canvas $C_t$, multi-modal synthesis is achieved by sampling a random image as the identity input ($y_t$) to the identity encoder network. Results are shown in Fig.~\ref{fig:multi-modal}. We observe that the above approach forms a convenient method for predicting multiple possible image completions from incomplete paintings. This provides the user with a wider range of choices to select the best direction for the synthesis process. Furthermore, we note that this idea can also be used to perform identity conditioned synthesis, by using the same identity image (\eg, Chris Hemsworth) throughout the painting trajectory.

\section{Ablation Study}

In this section, we perform several ablation studies in order to study the importance of different losses $\{\mathcal{L}_{pred},\mathcal{L}_{edit},\mathcal{L}_{embed}\}$ in the performance of \emph{paint2pix}. Please note that in order to still get meaningful results, the experiments without $\mathcal{L}_{pred}$ are performed while using a pretrained restyle \cite{alaluf2021restyle} network for independently predicting intermediate outputs $\{y_{t+1},y_{t}\}$ from canvas frames $\{C_{t+1},C_{t}\}$. 

\begin{figure}[h!]
\vskip -0.2in
\begin{center}
\centerline{\includegraphics[width=0.69\linewidth]{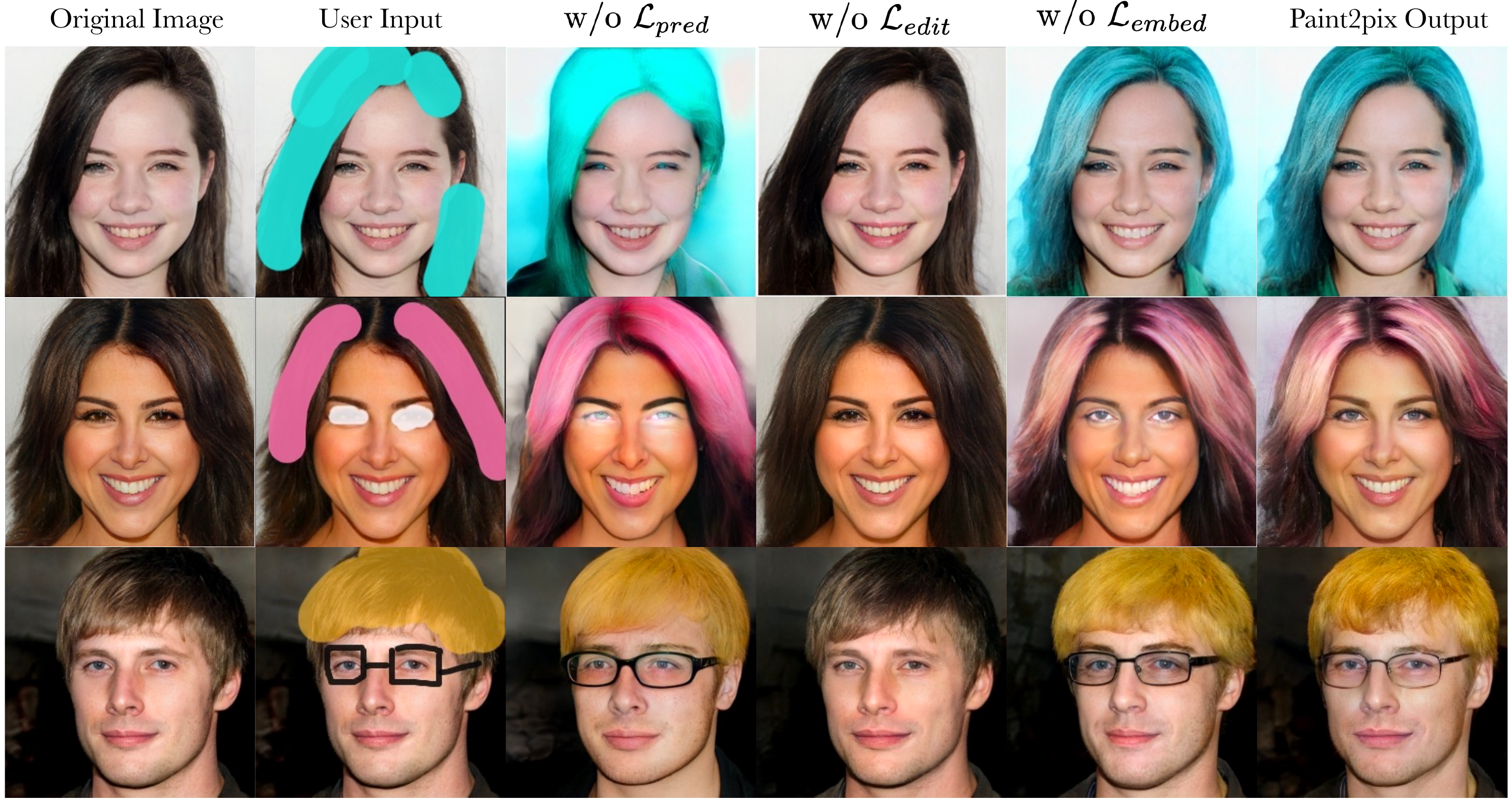}}
\vskip -0.1in
\caption{Ablation study for different losses in Paint2pix. Best viewed zoomed-in.}
\label{fig:ablation-studies}
\end{center}
\vskip -0.4in
\end{figure}

Results are shown in Fig.~\ref{fig:ablation-studies}. We observe that $\{\mathrm{w/o} \ \mathcal{L}_{pred}\}$ the model lacks an understanding of the manifold of incomplete paintings and thus produces outputs which are not fully photorealistic. In contrast, $\{\mathrm{w/o} \ \mathcal{L}_{edit}\}$  shows high quality outputs but does not incorporate the edits made by user brushstrokes. Finally, we see that the use of $\mathcal{L}_{embed}$ helps the model produce images which preserve the identity of the original image in the final prediction.

\newpage
\section{Discussion and Limitations}

We next provide a discussion of some advanced usage and limitations for \emph{paint2pix}, in order to aid a more holistic understanding of the proposed approach.
\vskip 0.05in
\noindent \textbf{In-distribution predictions.} A key advantage of \paint is that allows a novice user to synthesize and manipulate an output image on the real image manifold, while using fairly rudimentary and crude brushstrokes. While this is desirable in most scenarios, it also limits our method as it prevents a potential user from intentionally performing out-of-distribution (or non-realistic) facial manipulations (\eg, blue eyebrows, ghost like faces \etc).
\vskip 0.05in
\noindent \textbf{Invertibility for real-image editing.} Much like other GAN-inversion and latent space manipulation methods \cite{richardson2021encoding, alaluf2021restyle, zhu2016generative, patashnik2021styleclip,abdal2021styleflow,harkonen2020ganspace}, 
accurate real-image editing with \paint is highly dependent on the ability of used encoder architecture to invert the original real image into StyleGAN \cite{karras2020analyzing} latent space. 

\begin{figure}[h!]
\vskip -0.25in
\begin{center}
\subfigure[{\small Age Variation Edits}]{\includegraphics[width=0.5755\linewidth]{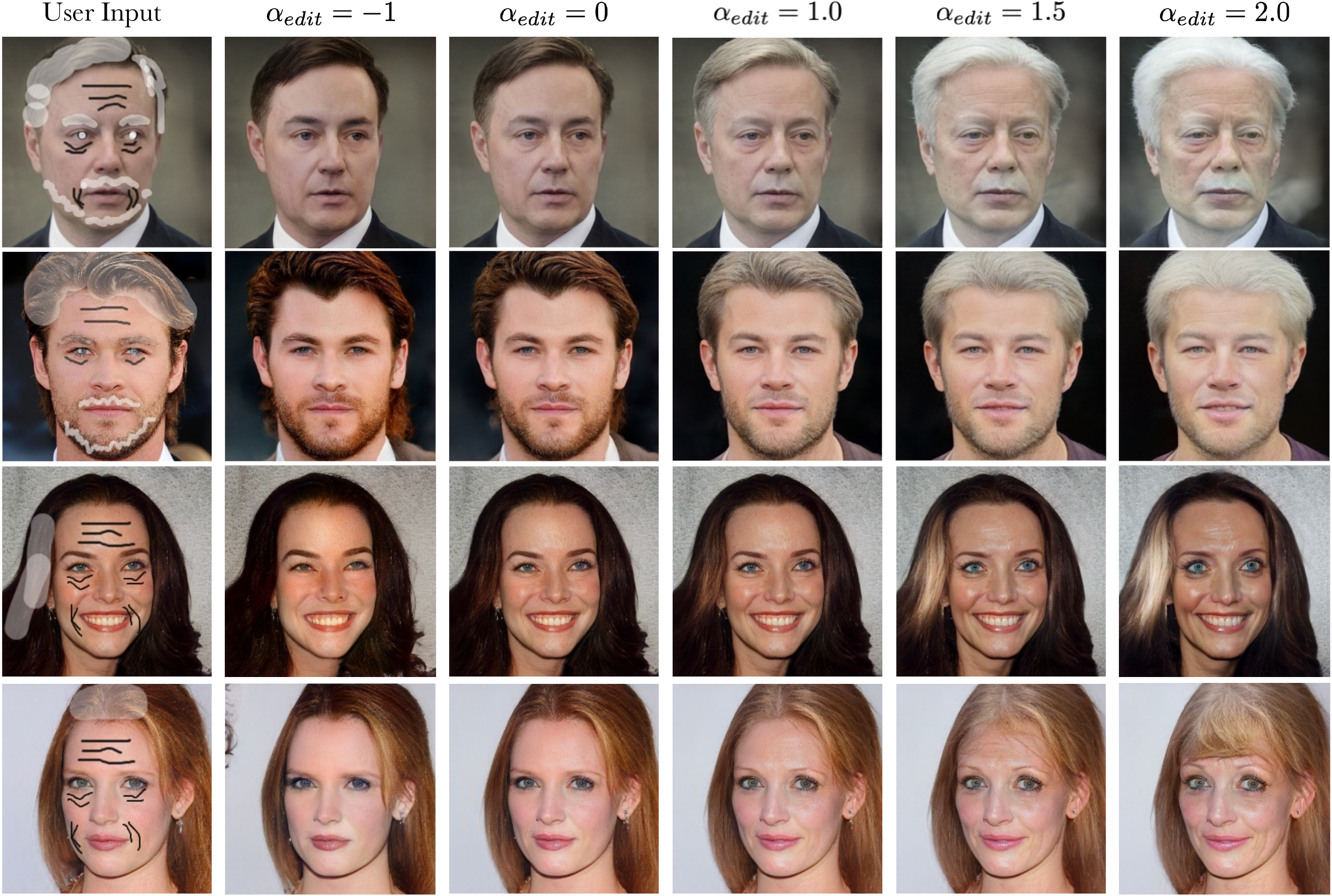}} 
\hfill
\subfigure[Gender Variation Edits]{\includegraphics[width=0.384\linewidth]{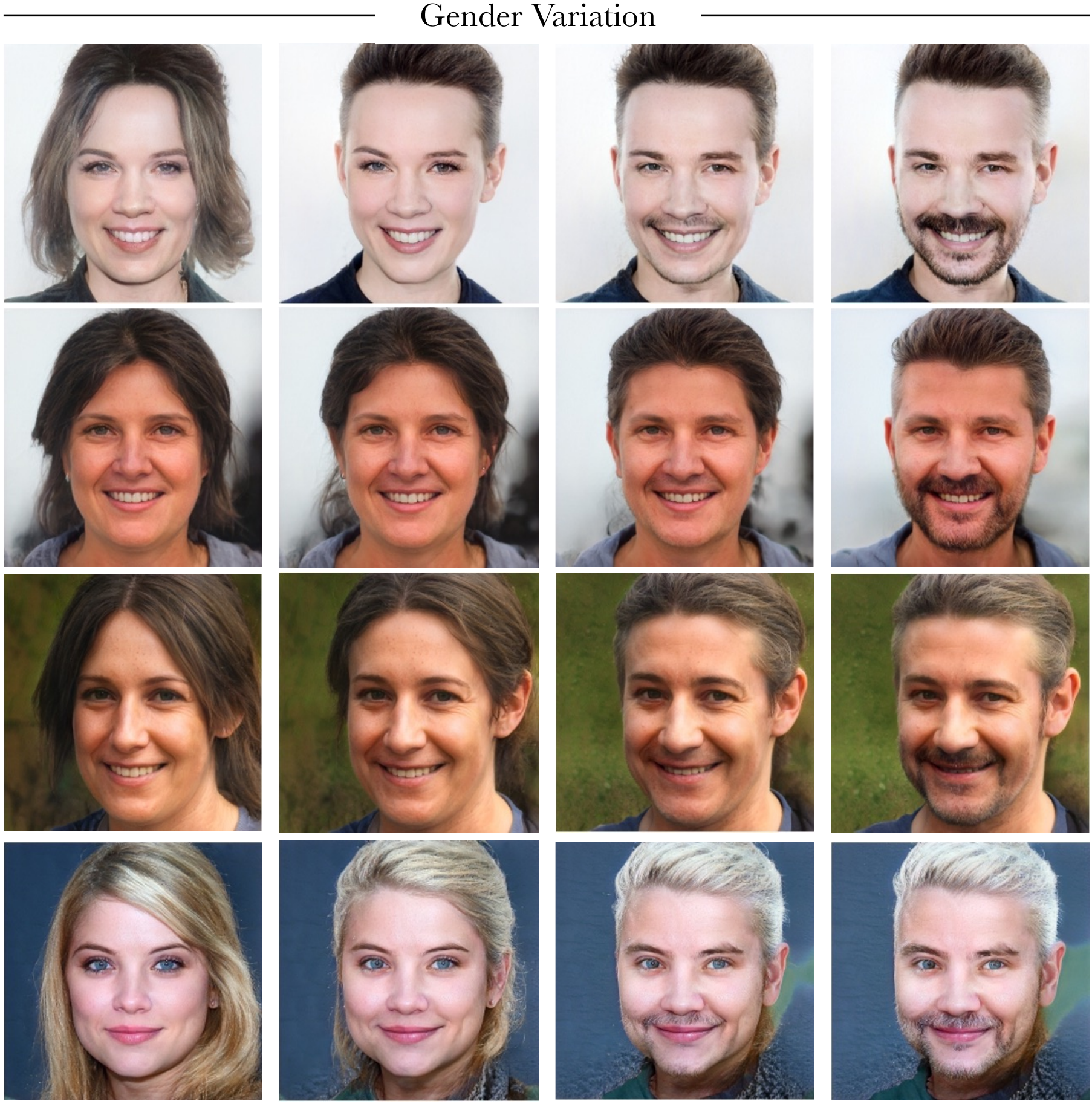}} 
\vskip -0.1in
\caption{Analysing paint2pix usage for achieving advanced semantic edits.}
\label{fig:discussion}
\end{center}
\vskip -0.5in
\end{figure}

\vskip 0.05in
\noindent \textbf{Advanced edits.} Another limitation is that \paint does not provide a direct approach for achieving advanced semantic edits like age, gender manipulation. Nevertheless, as show in Fig.~\ref{fig:discussion}, age variation edits can still be achieved using extrapolation of edit strength $\alpha$. Similarly, gender variation edits are possible by using progressive synthesis to infer the gender edit direction. Further details and analysis for gender variation edits are provided in the supp.~material.

\section{Conclusion}

In this paper, we explore a novel task of performing photorealistic image synthesis and editing using primitive user paintings and brushstrokes. To this end, we propose \paint which can be used for 1) progressively synthesizing a desired image output from scratch
using just few rudimentary brushstrokes,
or, 2) real image editing: wherein it allows a human user to directly perform a range of custom edits without requiring any artistic expertise. As shown through extensive experimentation, we find that \paint forms a highly convenient and simple approach for directly expressing a potential user's inner ideas in visual form.

\appendix

\clearpage
%
%
\bibliographystyle{splncs04}
\bibliography{egbib}

\end{document}